\documentclass[fleqn,10pt]{wlscirep}
\usepackage[utf8]{inputenc}
\usepackage[T1]{fontenc}

\usepackage{times}
\usepackage{hyperref}       
\usepackage{url}            
\usepackage{booktabs}       
\usepackage{amsfonts}       
\usepackage{nicefrac}       
\usepackage{microtype}      
\usepackage{lipsum}
\usepackage{float}
\usepackage{multicol}
\usepackage{multirow}
\usepackage{tabularx}
\usepackage{graphicx}
\usepackage{xcolor}
\usepackage{colortbl}
\usepackage{authblk}

\usepackage{amsmath}
\usepackage{amsthm}
\usepackage{enumerate}
\usepackage{enumitem}

\setlist{leftmargin=5.5mm}
\hypersetup{
    colorlinks=true,
    linkcolor=black,
    filecolor=blue,      
    urlcolor=blue,
}
\usepackage[textwidth=0.1in,textsize=footnotesize]{todonotes}
\usepackage{xspace}

\newcommand{\method}{PAMNet\xspace}

\newcommand{\angstrom}{\text{\normalfont\AA}}

\usepackage{lineno}

\title{A Universal Framework for Accurate and Efficient Geometric Deep Learning of Molecular Systems}

\author[1, 2]{Shuo Zhang}
\author[1]{Yang Liu}
\author[1,2,3,*]{Lei Xie}

\affil[1]{Department of Computer Science, Hunter College, The City University of New York, New York, 10065, USA}
\affil[2]{Helen \& Robert Appel Alzheimer’s Disease Research Institute, Feil Family Brain \& Mind Research Institute, Weill Cornell Medicine, Cornell University, New York, 10065, USA}
\affil[3]{Ph.D. Program in Computer Science, The Graduate Center, The City University of New York, New York, 10016, USA}
\affil[*]{lei.xie@hunter.cuny.edu}

\begin{abstract}
Molecular sciences address a wide range of problems involving molecules of different types and sizes and their complexes. Recently, geometric deep learning, especially Graph Neural Networks (GNNs), has shown promising performance in molecular science applications. However, most existing works often impose targeted inductive biases to a specific molecular system, and are inefficient when applied to macromolecules or large-scale tasks, thereby limiting their applications to many real-world problems. To address these challenges, we present PAMNet, a universal framework for accurately and efficiently learning the representations of three-dimensional (3D) molecules of varying sizes and types in any molecular system. Inspired by molecular mechanics, PAMNet induces a physics-informed bias to explicitly model local and non-local interactions and their combined effects. As a result, PAMNet can reduce expensive operations, making it time and memory efficient. In extensive benchmark studies, PAMNet outperforms state-of-the-art baselines regarding both accuracy and efficiency in three diverse learning tasks: small molecule properties, RNA 3D structures, and protein-ligand binding affinities. Our results highlight the potential for PAMNet in a broad range of molecular science applications.
\end{abstract}
\begin{document}

\flushbottom
\maketitle

\thispagestyle{empty}

\section*{Introduction}
The wide variety of molecular types and sizes poses numerous challenges in the computational modeling of molecular systems for drug discovery, structural biology, quantum chemistry, and others~\cite{holtje2003molecular}. To address these challenges, recent advances in geometric deep learning (GDL) approaches have become increasingly important~\cite{atz2021geometric,isert2022structure}. Especially, Graph Neural Networks (GNNs) have demonstrated superior performance among various GDL approaches~\cite{sun2020graph,bronstein2021geometric,reiser2022graph}. GNNs treat each molecule as a graph and perform message passing scheme on it~\cite{gilmer2017neural}. By representing atoms or groups of atoms like functional groups as nodes, and chemical bonds or any pairwise interactions as edges, molecular graphs can naturally encode the structural information in molecules. In addition to this, GNNs can incorporate symmetry and achieve invariance or equivariance to transformations such as rotations, translations, and reflections~\cite{han2022geometrically}, which further contributes to their effectiveness in molecular science applications. To enhance their ability to capture molecular structures and increase the expressive power of their models, previous GNNs have utilized auxiliary information such as chemical properties~\cite{duvenaud2015convolutional,kearnes2016molecular,chen2019graph,yang2019analyzing}, atomic pairwise distances in Euclidean space~\cite{gilmer2017neural,schutt2018schnetpack,unke2019physnet}, angular information~\cite{klicpera_dimenet_2020,klicpera_dimenetpp_2020,shui2020heterogeneous,li2021structure}, etc. 

In spite of the success of GNNs, their application in molecular sciences is still in its early stages. One reason for this is that current GNNs often use targeted inductive bias for modeling a specific type of molecular system, and cannot be directly transferred to other contexts although all molecule structures and their interactions follow the same law of physics. For example, GNNs designed for modeling proteins may include operations that are specific to the structural characteristics of amino acids~\cite{jumper2021highly,baek2021accurate}, which are not relevant for other types of molecules.
Additionally, GNNs that incorporate comprehensive geometric information can be computationally expensive, making them difficult to scale to tasks involving a large number of molecules (e.g., high-throughput compound screening) or macromolecules (e.g., proteins and RNAs). For instance, incorporating angular information can significantly improve the performance of GNNs~\cite{klicpera_dimenet_2020,klicpera_dimenetpp_2020,shui2020heterogeneous,li2021structure}, but also increases the complexity of the model, requiring at least $O(Nk^2)$ messages to be computed where $N$ and $k$ denote the number of nodes and the average degree in a graph.

To tackle the limitations mentioned above, we propose a universal GNN framework, \underline{P}hysics-\underline{A}ware \underline{M}ultiplex Graph Neural \underline{Net}work (\method), for the accurate and efficient representation learning of 3D molecules ranging from small molecules to macromolecules in any molecular system. \method induces a physics-informed bias inspired by molecular mechanics~\cite{schlick2010molecular}, which separately models local and non-local interactions in molecules based on different geometric information. To achieve this, we represent each molecule as a two-layer multiplex graph, where one plex only contains local interactions, and the other plex contains additional non-local interactions. \method takes the multiplex graphs as input and uses different operations to incorporate the geometric information for each type of interaction. This flexibility allows \method to achieve efficiency by avoiding the use of computationally expensive operations on non-local interactions, which consist of the majority of interactions in a molecule. Additionally, a fusion module in \method allows the contribution of each type of interaction to be learned and fused for the final feature or prediction. To preserve symmetry, \method utilizes E(3)-invariant representations and operations when predicting scalar properties, and is extended to predict E(3)-equivariant vectorial properties by considering the geometric vectors in molecular structures that arise from quantum mechanics.

To demonstrate the effectiveness of \method, we conduct a comprehensive set of experiments on a variety of tasks involving different molecular systems, including small molecules, RNAs, and protein-ligand complexes. These tasks include predicting small molecule properties, RNA 3D structures, and protein-ligand binding affinities. We compare \method to state-of-the-art baselines in each task and the results show that \method outperforms the baselines in terms of both accuracy and efficiency across all three tasks. Given the diversity of the tasks and the types of molecules involved, the superior performance of \method shows its versatility to be applied in various real-world scenarios.

\begin{figure}[t]
    \centering
    \includegraphics[width=1.0\textwidth]{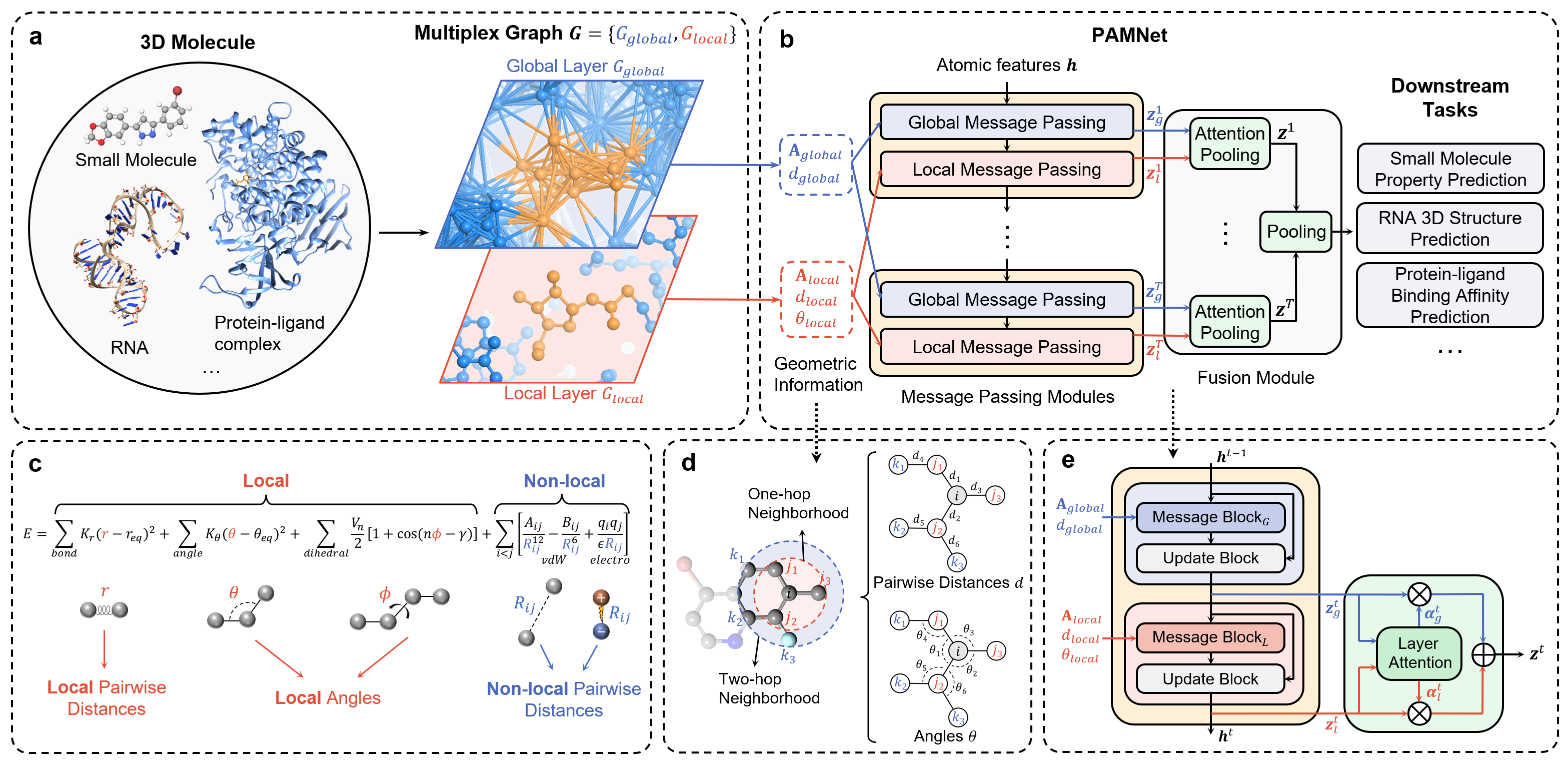}
    \caption{\label{fig:method}\textbf{Overview of \method.} \textbf{a}, Based on the 3D structure of any molecule or molecular system, a two-layer multiplex graph $G = \{G_{global}, G_{local}\}$ is constructed to separate the modeling of global and local interactions. \textbf{b}, \method takes $G$ as input and learns node-level or graph-level representation for downstream tasks. \method contains stacked message passing modules that update the node embeddings $\textbf{z}$ in $G$, and a fusion module that learns to combine the updated embeddings. In each message passing module, two message passing schemes are designed to encode the different geometric information in $G$'s two layers. In the fusion module, a two-step pooling process is proposed. \textbf{c}, Calculation of molecular energy $E$ in molecular mechanics. \textbf{d}, An example of the geometric information in $G$. By considering the one-hop neighbors $\{j\}$ and two-hop neighbors $\{k\}$ of atom $i$, we can define the pairwise distances $d$ and the related angles $\theta$. \textbf{e}, Detailed architecture of the message passing module and the attention pooling in \method.}
\end{figure}

\section*{Overview of \method}
\paragraph{Multiplex graph representation.}
Given any 3D molecule or molecular system, we define a multiplex graph representation as the input of our \method model based on the original 3D structure (Fig.~\ref{fig:method}a). The construction of multiplex graphs is inspired by molecular mechanics~\cite{schlick2010molecular}, in which the molecular energy $E$ is separately modeled based on local and non-local interactions (Fig.~\ref{fig:method}c). \textcolor{black}{In detail, the local terms $E_{\text {bond}}+E_{\text {angle}}+E_{\text {dihedral}}$ model local, covalent interactions including $E_{\text {bond}}$ that depends on bond lengths, $E_{\text {angle}}$ on bond angles, and $E_{\text {dihedral}}$ on dihedral angles. The non-local terms $E_{\text {vdW}}+E_{\text {electro}}$ model non-local, non-covalent interactions including van der Waals and electrostatic interactions which depend on interatomic distances.} Motivated by this, we also decouple the modeling of these two types of interactions in \method. For local interactions, we can define them either using chemical bonds or by finding the neighbors of each node within a relatively small cutoff distance, depending on the given task. For global interactions that contain both local and non-local ones, we define them by finding the neighbors of each node within a relatively large cutoff distance. For each type of interaction, we use a layer to represent all atoms as nodes and the interactions as edges. The resulting layers that share the same group of atoms form a two-layer multiplex graph $G = \{G_{global}, G_{local}\}$ which represents the original 3D molecular structure (Fig.~\ref{fig:method}a).

\paragraph{Message passing modules.}
To update the node embeddings in the multiplex graph $G$, we design two message passing modules that incorporate geometric information: \textit{Global Message Passing} and \textit{Local Message Passing} for updating the node embeddings in $G_{global}$ and $G_{local}$, respectively (Fig.~\ref{fig:method}b). These message passing modules are inspired by physical principles from molecular mechanics (Fig.~\ref{fig:method}c): When modeling the molecular energy $E$, the terms for local interactions require geometric information including interatomic distances (bond lengths) and angles (bond angles and dihedral angles), while the terms for non-local interactions only require interatomic distances as geometric information. The message passing modules in \method also use geometric information in this way when modeling these interactions (Fig.~\ref{fig:method}b and Fig.~\ref{fig:method}e). Specifically, we capture the pairwise distances and angles contained within up to two-hop neighborhoods (Fig.~\ref{fig:method}d). The \textit{Local Message Passing} requires the related adjacency matrix $\textbf{A}_{local}$, pairwise distances $d_{local}$ and angles $\theta_{local}$, while the \textit{Global Message Passing} only needs the related adjacency matrix $\textbf{A}_{global}$ and pairwise distances $d_{global}$. Each message passing module then learns the node embeddings $\textbf{z}_g$ or $\textbf{z}_l$ in $G_{global}$ and $G_{local}$, respectively.

\textcolor{black}{For the operations in our message passing modules, they can preserve different symmetries: E(3)-invariance and E(3)-equivariance, which contain essential inductive bias incorporated by GNNs when dealing with graphs with geometric information~\cite{han2022geometrically}. E(3)-invariance is preserved when predicting E(3)-invariant scalar quantities like energies, which remain unchanged when the original molecular structure undergoes any E(3) transformation including rotation, translation, and reflection. To preserve E(3)-invariance,} the input node embeddings $\textbf{\textit{h}}$ and geometric features are all E(3)-invariant. To update these features, \method utilizes operations that can preserve the invariance. \textcolor{black}{In contrast, E(3)-equivariance is preserved when predicting E(3)-equivariant vectorial quantities like dipole moment, which will change according to the same transformation applied to the original molecular structure through E(3) transformation. To preserve E(3)-equivariance,} an extra associated geometric vector $\vec{v} \in \mathbb{R}^3$ is defined for each node. These geometric vectors are updated by operations inspired by quantum mechanics~\cite{stone1981distributed}, allowing for the learning of E(3)-equivariant vectorial representations. More details about \textcolor{black}{the explanations of E(3)-invariance, E(3)-equivariance, and our} operations can be found in Methods.

\paragraph{Fusion module.}
After updating the node embeddings $\textbf{z}_g$ or $\textbf{z}_l$ of the two layers in the multiplex graph $G$, we design a fusion module with a two-step pooling process to combine $\textbf{z}_g$ and $\textbf{z}_l$ for downstream tasks (Figure~\ref{fig:method}b). In the first step, we design an attention pooling module based on attention mechanism~\cite{velivckovic2018graph} for each hidden layer $t$ in \method. Since $G_{global}$ and $G_{local}$ contains the same set of nodes $\{\textit{N}\}$, we apply the attention mechanism to each node $n \in \{\textit{N}\}$ to learn the attention weights ($\alpha_g^t$ and $\alpha_l^t$) between the node embeddings of $n$ in $G_{global}$ and $G_{local}$, which are $\textbf{z}_g^t$ and $\textbf{z}_l^t$. Then the attention weights are treated as the importance of $\textbf{z}_g^t$ and $\textbf{z}_l^t$ to compute the combined node embedding $\textbf{z}^t$ in each hidden layer $t$ based on a weighted summation (Figure~\ref{fig:method}e). In the second step, the $\textbf{z}^t$ of all hidden layers are summed together to compute the node embeddings of the original input. If a graph embedding is desired, we compute it using an average or a summation of the node embeddings. 

\section*{Results and discussion}
In this section, we will demonstrate the performance of our proposed \method regarding two aspects: accuracy and efficiency. Accuracy denotes how well the model performs measured by the metrics corresponding to a given task. Efficiency denotes the memory consumed and the inference time spent by the model.

\subsection*{Performance of \method regarding accuracy}
\paragraph{Small molecule property prediction.}
To evaluate the accuracy of \method in learning representations of small 3D molecules, we choose QM9, which is a widely used benchmark for the prediction of 12 molecular properties of around 130k small organic molecules with up to 9 non-hydrogen atoms~\cite{ramakrishnan2014quantum}. Mean absolute error (MAE) and mean standardized MAE (std. MAE)~\cite{klicpera_dimenet_2020} are used for quantitative evaluation of the target properties. Besides evaluating the original \method which captures geometric information within two-hop neighborhoods of each node, we also develop a "simple" \method, called \method-s, that utilizes only the geometric information within one-hop neighborhoods. The \method models are compared with several state-of-the-art models including SchNet~\cite{schutt2018schnetpack}, PhysNet~\cite{unke2019physnet}, MGCN~\cite{lu2019molecular}, PaiNN~\cite{schutt2021equivariant}, DimeNet++~\cite{klicpera_dimenetpp_2020}, and SphereNet~\cite{liu2022spherical}. More details of the experiments can be found in Methods and Supplementary Information. 

We compare the performance of \method with those of the baseline models mentioned above on QM9, as shown in Table~\ref{table:QM9}. \method achieves 4 best and 6 second-best results among all 12 properties, while \method-s achieves 3 second-best results. When evaluating the overall performance using the std. MAE across all properties, \method and \method-s rank 1 and 2 among all models with 10$\%$ and 5$\%$ better std. MAE than the third-best model (SphereNet), respectively. From the results, we can observe that the models incorporating only atomic pairwise distance $d$ as geometric information like SchNet, PhysNet, and MGCN generally perform worse than those models incorporating more geometric information like PaiNN, DimeNet++, SphereNet, and our \method. Besides, \method-s which captures geometric information only within one-hop neighborhoods performs worse than \method which considers two-hop neighborhoods. These show the importance of capturing rich geometric information when representing 3D small molecules. The superior performance of \method models demonstrates the power of our separate modeling of different interactions in molecules and the effectiveness of the message passing modules designed.

When predicting dipole moment $\mu$ as a scalar value, which is originally an E(3)-equivariant vectorial property $\vec{\mu}$, \method preserves the E(3)-equivariance to directly predict $\vec{\mu}$ first and then takes the magnitude of $\vec{\mu}$ as the final prediction. As a result, \method and \method-s all get lower MAE (10.8 mD and 11.3 mD) than the previous best result (12 mD) achieved by PaiNN, which is a GNN with equivariant operations for predicting vectorial properties. Note that the remaining baselines all directly predict dipole moment as a scalar property by preserving invariance. We also examine that by preserving invariance in \method and directly predicting dipole moment as a scalar property, the MAE (24.0 mD) is much higher than the equivariant version. These results demonstrate that preserving equivariance is more helpful than preserving invariance for predicting dipole moments.

\begin{table}[t]
\centering
\small
\begin{tabular}{lccccccccc}
\toprule
    {Property} & {Unit} & {SchNet} & {PhysNet} & {MGCN} & {PaiNN} & {DimeNet++} & {SphereNet} & \textbf{\method-s}  & \textbf{\method}\\
\midrule
    $\mu$  &  mD     & 21 & 52.9 & 56 & 12 & 29.7 & 24.5 & \underline{11.3} & \textbf{10.8}\\
    $\alpha$ & $a_0^3$  & 0.124 & 0.0615 & \textbf{0.030} & 0.045 & \underline{0.0435} & 0.0449 & 0.0466 & 0.0447\\
    $\epsilon_{\text{HOMO}}$ & meV  & 47 & 32.9 & 42.1 & 27.6 & 24.6 & \textbf{22.8} & \underline{23.9}  & \textbf{22.8}\\
    $\epsilon_{\text{LUMO}}$ & meV  & 39 & 24.7 & 57.4 & 20.4 & 19.5 & \textbf{18.9} & 20.0  & \underline{19.2}\\
    $\Delta\epsilon$ & meV    & 74 & 42.5 & 64.2 & 45.7 & 32.6 & \underline{31.1} & 32.4 & \textbf{31.0}\\
    $\left\langle R^{2}\right\rangle$ & $a_0^2$  & 0.158 & 0.765 & 0.11 & \textbf{0.066} & 0.331 & 0.268 & 0.094 & \underline{0.093}\\
    ZPVE & meV  & 1.616 & 1.39 & \textbf{1.12} & 1.28 & 1.21 & \textbf{1.12} & 1.24 & \underline{1.17}\\
    $U_0$ & meV & 12 & 8.15 & 12.9 & \textbf{5.85} & 6.32 & 6.26 & 6.05 & \underline{5.90}\\
    $U$ & meV  & 12 & 8.34 & 14.4 & \textbf{5.83} & 6.28 & 6.36 & 6.08 & \underline{5.92} \\
    $H$ & meV   & 12 & 8.42 & 16.2 & \textbf{5.98} & 6.53 & 6.33 & 6.19 & \underline{6.04} \\
    $G$ & meV   & 13 & 9.40 & 14.6 & 7.35 & 7.56 & 7.78 & \underline{7.34} & \textbf{7.14} \\
    $c_v$ & $\frac{\mathrm{cal}}{\mathrm{mol} \mathrm{K}}$  & 0.034 & 0.0280 & 0.038 & 0.024 & \underline{0.0230} & \textbf{0.0215} & 0.0234 & 0.0231 \\
\midrule
    std. MAE & $\%$  & 1.78 & 1.37 & 1.89 & 1.01 & 0.98 & 0.91 & \underline{0.87} & \textbf{0.83}\\
\bottomrule
\end{tabular}
\caption{\textbf{Performance comparison on QM9.} The best results are marked in bold and the second-best results with underline.}
\label{table:QM9}
\end{table}

\begin{figure}[t]
    \centering
    \includegraphics[width=0.75\columnwidth]{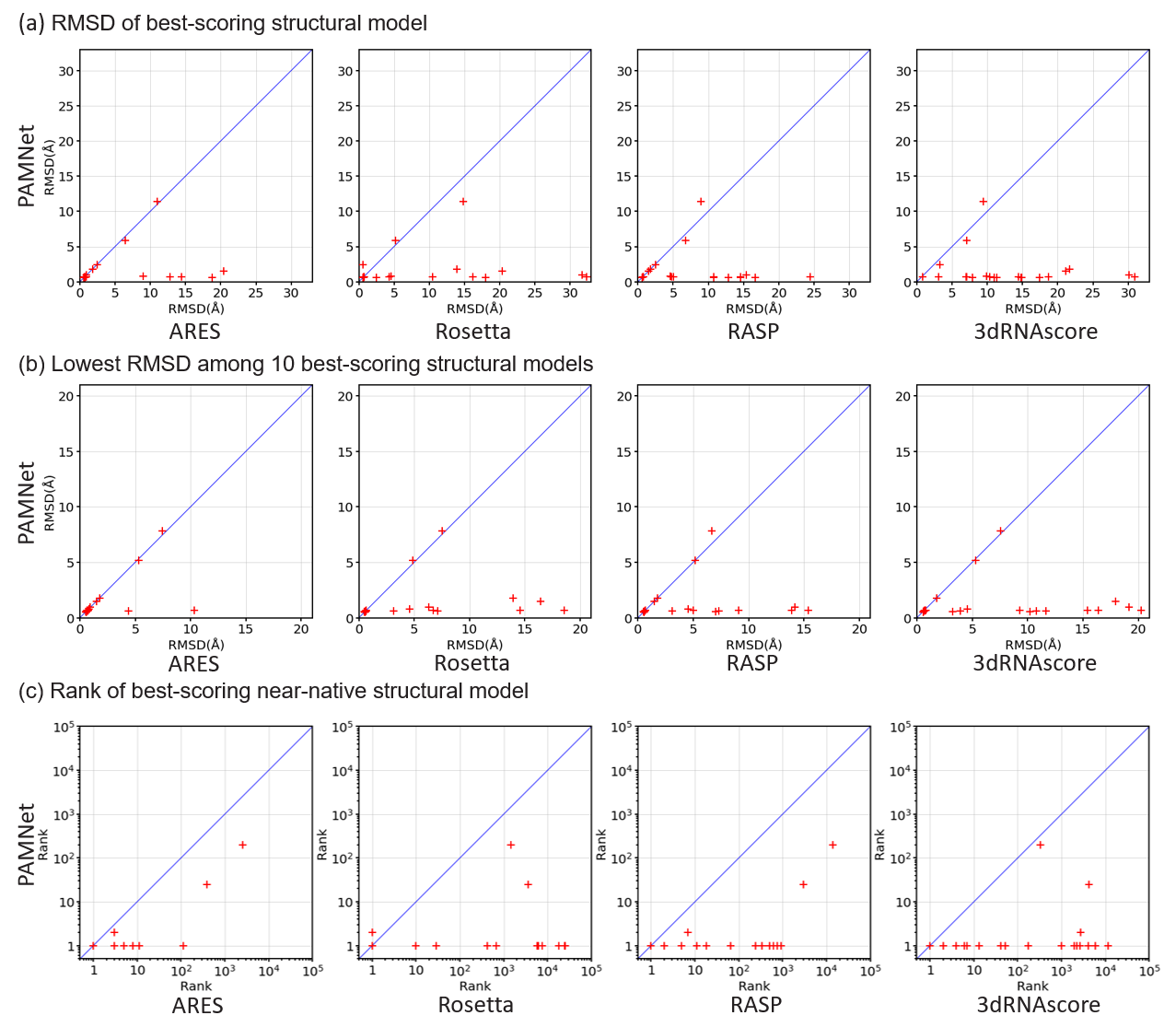}
    \caption{\label{fig:benchmark}\textbf{Performance comparison on RNA-Puzzles.} Given a group of candidate structural models for each RNA, we rank the models using \method and the other four leading scoring functions for comparison. Each cross in the figures corresponds to one RNA. (a) The best-scoring structural model of each RNA predicted by the scoring functions is compared. \method in general identifies more accurate models (with lower RMSDs from the native structure) than those decided by the other scoring functions. (b) Comparison of the 10 best-scoring structural models. The identifications of \method contain accurate models more frequently than those from other scoring functions. (c) The rank of the best-scoring near-native structural model for each RNA is used for comparison. \method usually performs better than the other scoring functions by having a lower rank.}
\end{figure}

\paragraph{RNA 3D structure prediction.}
Besides small molecules, we further apply \method to predict RNA 3D structures for evaluating the accuracy of \method in learning representations of 3D macromolecules. Following the previous works~\cite{wang20153drnascore,watkins2020farfar2,townshend2021geometric}, we refer the prediction to be the task of identifying accurate structural models of RNA from less accurate ones: Given a group of candidate 3D structural models generated based on an RNA sequence, a desired model that serves as a scoring function needs to distinguish accurate structural models among all candidates. We use the same datasets as those used in~\cite{townshend2021geometric}, which include a dataset for training and a benchmark for evaluation. The training dataset contains 18 relatively older and smaller RNA molecules experimentally determined~\cite{das2007automated}. Each RNA is used to generate 1000 structural models via the Rosetta FARFAR2 sampling method~\cite{watkins2020farfar2}. The benchmark for evaluation contains relatively newer and larger RNAs, which are the first 21 RNAs in the RNA-Puzzles structure prediction challenge~\cite{miao2020rna}. Each RNA is used to generate at least 1500 structural models using FARFAR2, where 1$\%$ of the models are near-native (i.e., within a 2$\angstrom$ RMSD of the experimentally determined native structure). In practice, each scoring function predicts the root mean square deviation (RMSD) from the unknown true structure for each structural model. A lower RMSD would suggest a more accurate structural model predicted. We compare \method with four state-of-the-art baselines: ARES~\cite{townshend2021geometric}, Rosetta (2020 version)~\cite{watkins2020farfar2}, RASP~\cite{capriotti2011all}, and 3dRNAscore~\cite{wang20153drnascore}. Among the baselines, only ARES is a deep learning-based method, and is a GNN using equivariant operations. More details of the experiments are introduced in Methods and Supplementary Information.

On the RNA-Puzzles benchmark for evaluation, \method significantly outperforms all other four scoring functions as shown in Figure~\ref{fig:benchmark}. When comparing the best-scoring structural model of each RNA (Figure~\ref{fig:benchmark}a), the probability of the model to be near-native (<2$\angstrom$ RMSD from the native structure) is 90$\%$ when using \method, compared with 62, 43, 33, and 5$\%$ for ARES, Rosetta, RASP, and 3dRNAscore, respectively. As for the 10 best-scoring structural models of each RNA (Figure~\ref{fig:benchmark}b), the probability of the models to include at least one near-native model is 90$\%$ when using \method, compared with 81, 48, 48, and 33$\%$ for ARES, Rosetta, RASP, and 3dRNAscore, respectively. When comparing the rank of the best-scoring near-native structural model of each RNA (Figure~\ref{fig:benchmark}c), the geometric mean of the ranks across all RNAs is 1.7 for \method, compared with 3.6, 73.0, 26.4, and 127.7 for ARES, Rosetta, RASP, and 3dRNAscore, respectively. The lower mean rank of \method indicates that less effort is needed to go down the ranked list of \method to include one near-native structural model. A more detailed analysis of the near-native ranking task can be found in Supplementary Figure~\ref{fig:rna_appendix}.

\begin{table}[t]
\small
\centering
\begin{tabular}{clcccc}
\toprule
    \multicolumn{2}{c}{Model} & RMSE $\downarrow$ & MAE $\downarrow$ & SD $\downarrow$ & R $\uparrow$ \\
\midrule
     \multirow{3}{*}{ML-based} & LR & 1.675 (0.000)  & 1.358 (0.000)  & 1.612 (0.000)  & 0.671 (0.000) \\
     & SVR & 1.555 (0.000)  & 1.264 (0.000)  & 1.493 (0.000)  & 0.727 (0.000) \\
     & RF-Score & 1.446 (0.008)  & 1.161 (0.007)  & 1.335 (0.010)  & 0.789 (0.003) \\
\midrule
     \multirow{2}{*}{CNN-based} & Pafnucy & 1.585 (0.013)  & 1.284 (0.021)  & 1.563 (0.022)  & 0.695 (0.011) \\
     & OnionNet & 1.407 (0.034)  & 1.078 (0.028)  & 1.391 (0.038)  & 0.768 (0.014) \\
\midrule
     \multirow{8}{*}{GNN-based} & GraphDTA & 1.562 (0.022)  & 1.191 (0.016)  & 1.558 (0.018)  & 0.697 (0.008) \\
     & SGCN & 1.583 (0.033)  & 1.250 (0.036)  & 1.582 (0.320)  & 0.686 (0.015) \\
     & GNN-DTI & 1.492 (0.025)  & 1.192 (0.032)  & 1.471 (0.051)  & 0.736 (0.021) \\
     & D-MPNN & 1.493 (0.016) & 1.188 (0.009) & 1.489 (0.014) & 0.729 (0.006) \\
     & MAT & 1.457 (0.037)  & 1.154 (0.037)  & 1.445 (0.033)  & 0.747 (0.013) \\
     & DimeNet & 1.453 (0.027) & 1.138 (0.026) & 1.434 (0.023) & 0.752 (0.010) \\
     & CMPNN & 1.408 (0.028) & 1.117 (0.031) & 1.399 (0.025) & 0.765 (0.009) \\
     & SIGN & \underline{1.316} (0.031) & \underline{1.027} (0.025) & \underline{1.312} (0.035) & \underline{0.797} (0.012)\\
\midrule
     Ours & \textbf{\method} & \textbf{1.263 (0.017)} & \textbf{0.987 (0.013)} & \textbf{1.261 (0.015)} & \textbf{0.815 (0.005)}\\
\bottomrule
\end{tabular}
\caption{\textbf{Performance comparison on PDBbind.} We report the averaged results together with standard deviations. For the evaluation metrics, $\downarrow$ denotes the lower the better, while $\uparrow$ denotes the higher the better. The best results are marked in bold and the second-best results with underline. }
\label{table:PDBBind}
\end{table}

\begin{table}[t]
\small
\centering
\begin{tabular}{llrr}
\toprule
    Dataset & Model & Memory (GB) & Inference Time (s) \\
\midrule
    \multirow{4}{*}{QM9} & DimeNet++ & 21.1 & 11.3 \\
     & SphereNet & 22.7 & 11.1 \\
     & \textbf{\method-s} & \textbf{6.0} & \textbf{7.3} \\
     & \textbf{\method} & 6.2 & 11.0 \\
\midrule
    \multirow{2}{*}{RNA-Puzzles} & ARES & 13.5 & 2.1 \\
     & \textbf{\method} & \textbf{7.8} & \textbf{0.6} \\
\midrule
    \multirow{2}{*}{PDBbind} & SIGN & 19.7 & 12.0 \\
     & \textbf{\method} & \textbf{13.1} & \textbf{1.8} \\
\bottomrule
\end{tabular}
\caption{\textbf{Results of efficiency evaluation.} We compare \method with the best-performed baselines in each of the three tasks regarding memory consumption and inference time. The most efficient results are marked in bold.}
\label{table:efficiency}
\end{table}

\begin{figure}[t]
    \centering
    \includegraphics[width=0.4 \columnwidth]{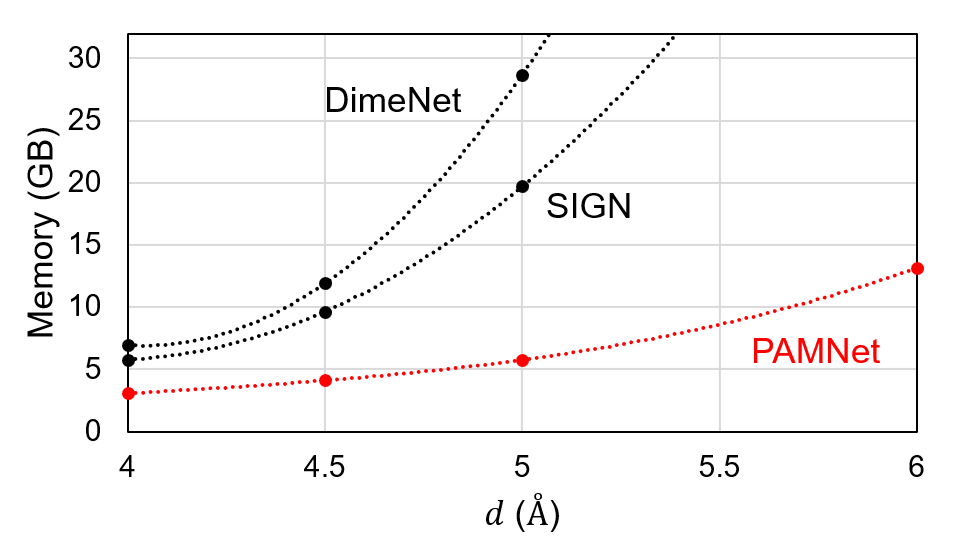}
    \caption{\textbf{Memory consumption vs. the largest cutoff distance $d$ on PDBbind.} We compare \method with the GNN baselines that also explicitly incorporate the 3D molecular geometric information like pairwise distances and angles.}
    \label{fig:efficiency}
\end{figure}

\begin{figure}[t]
    \centering
    \includegraphics[width=0.85 \columnwidth]{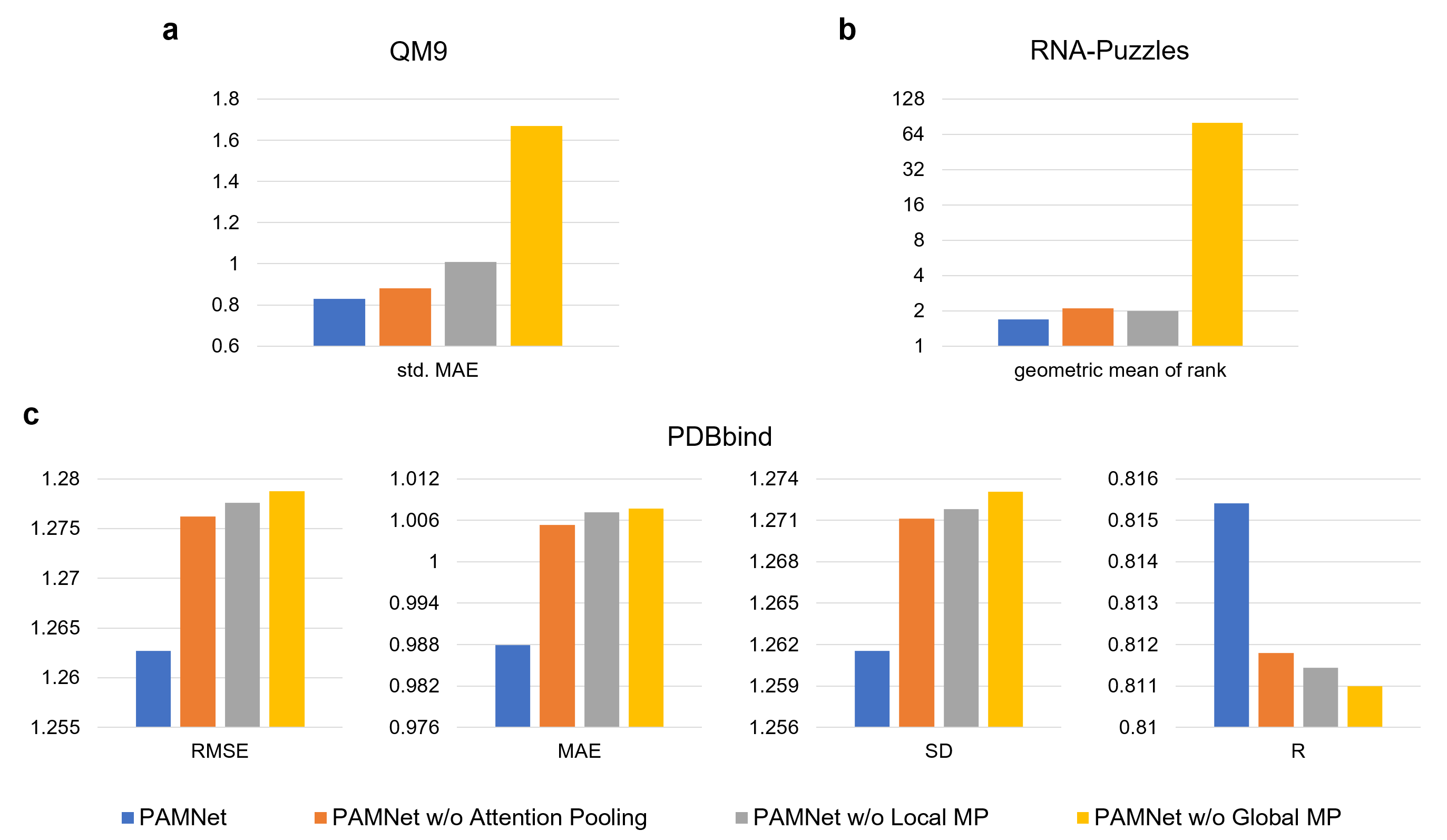}
    \caption{\textbf{Ablation study of \method}. We compare the variants with the original \method and report the differences.}
    \label{fig:ablation}
\end{figure}

\paragraph{Protein-ligand binding affinity prediction.}
In this experiment, we evaluate the accuracy of \method in representing the complexes that contain both small molecules and macromolecules. We use PDBbind, which is a well-known public database of experimentally measured binding affinities for protein-ligand complexes~\cite{wang2004pdbbind}. The goal is to predict the binding affinity of each complex based on its 3D structure. We use the PDBbind v2016 dataset and preprocess each original complex to a structure that contains around 300 nonhydrogen atoms on average with only the ligand and the protein residues within 6$\angstrom$ around it. To comprehensively evaluate the performance, we use Root Mean Square Error (RMSE), Mean Absolute Error (MAE), Pearson’s correlation coefficient (R) and the standard deviation (SD) in regression following~\cite{li2021structure}. \method is compared with various comparative methods including machine learning-based methods (LR, SVR, and RF-Score~\cite{ballester2010machine}), CNN-based methods (Pafnucy~\cite{stepniewska2018development} and OnionNet~\cite{zheng2019onionnet}), and GNN-based methods (GraphDTA~\cite{nguyen2021graphdta}, SGCN~\cite{danel2020spatial}, GNN-DTI~\cite{lim2019predicting}, D-MPNN~\cite{yang2019analyzing}, MAT~\cite{maziarka2020molecule},
DimeNet~\cite{klicpera_dimenet_2020}, CMPNN~\cite{song2020communicative}, and SIGN~\cite{li2021structure}). More details of the experiments are provided in Methods and Supplementary Information. 

We list the results of all models and compare their performance in Table~\ref{table:PDBBind} and Supplementary Table~\ref{table:significance}. \method achieves the best performance regarding all 4 evaluation metrics in our experiment. When compared with the second-best model, SIGN, our \method performs significantly better with p-value < 0.05. These results clearly demonstrate the accuracy of our model when learning representations of 3D macromolecule complexes. 

In general, we find that the models with explicitly encoded 3D geometric information like DimeNet, SIGN, and our \method outperform the other models without the information directly encoded. An exception is that DimeNet cannot beat CMPNN. This might be because DimeNet is domain-specific and is originally designed for small molecules rather than macromolecule complexes. In contrast, our proposed \method is more flexible to learn representations for various types of molecular systems. The superior performance of \method for predicting binding affinity relies on the separate modeling of local and non-local interactions. For protein-ligand complexes, the local interactions mainly capture the interactions inside the protein and the ligand, while the non-local interactions can capture the interactions between protein and ligand. Thus \method is able to effectively handle diverse interactions and achieve accurate results.

\subsection*{Performance of \method regarding efficiency}
To evaluate the efficiency of \method, we compare it to the best-performed baselines in each task regarding memory consumption and inference time and summarize the results in Table~\ref{table:efficiency}. Theoretically, DimeNet++, SphereNet, and SIGN all require $O(Nk^2)$ messages in message passing, while our \method requires $O(N(k_g+{k_l}^2))$ messages instead, where $N$ is the number of nodes, $k$ is the average degree in a graph, $k_g$ and $k_l$ denotes the average degree in $G_g$ and $G_l$ in the corresponding multiplex graph $G$. When $k_g \sim k$ and $k_l \ll k_g$, \method is much more efficient regarding the number of messages involved. A more detailed analysis of computational complexity is included in Methods. Based on the results in Table~\ref{table:efficiency} empirically, we find \method models all require less memory consumption and inference time than the best-performed baselines in all three tasks, which matches our theoretical analysis. We also compare the memory consumption when using a different largest cutoff distance $d$ of the related models in Figure~\ref{fig:efficiency}. From the results, we observe that the memory consumed by DimeNet and SIGN increases much faster than \method when $d$ increases. When fixing $d=5\angstrom$ as an example, \method requires 80$\%$ and 71$\%$ less memory than DimeNet and SIGN, respectively. Thus \method is much more memory-efficient and is able to capture longer-range interactions than these baselines with restricted resources. The efficiency of \method models comes from the separate modeling of local and non-local interactions in 3D molecular structures. By doing so, when modeling the non-local interactions, which make up the majority of all interactions, we utilize a relatively efficient message passing scheme that only encodes pairwise distances $d$ as the geometric information. Thus when compared with the models that require more comprehensive geometric information when modeling all interactions, \method significantly reduces the computationally expensive operations. More information about the details of experimental settings is included in Methods.

\subsection*{All components in \method contribute to the performance
}
To figure out whether all of the components in \method, including the fusion module and the message passing modules, contribute to the performance of \method, we conduct an ablation study by designing \method variants. Without the attention pooling, we use the averaged results from the message passing modules in each hidden layer to build a variant. We also remove either the Local Message Passing or the Global Message Passing for investigation. The performances of all \method variants are evaluated on the three benchmarks. Specifically, \textcolor{black}{the std. MAE across all properties} on QM9, the geometric mean of the ranks across all RNAs on RNA-Puzzles, and the four metrics used in the experiment on PDBbind are computed for comparison. The results in Figure~\ref{fig:ablation} show that all variants decrease the performance of \method in the evaluations, which clearly validates the contributions of all those components. \textcolor{black}{Detailed results of the properties on QM9 can be found in Supplementary Table~\ref{table:ablation_qm9}.}

\subsection*{Analysis of the contribution of local and global interactions}
A salient property of \method is the incorporation of the attention mechanism in the fusion module, which takes the importance of node embeddings in $G_{local}$ and $G_{global}$ of $G$ into consideration in learning combined node embeddings. Recall that for each node $n$ in the set of nodes $\{N\}$ in $G$, the attention pooling in the fusion module learns the attention weights $\alpha_l$ and $\alpha_g$ between $n$'s node embedding $\textbf{z}_{l}$ in $G_{local}$ and $n$'s node embedding $\textbf{z}_{g}$ in $G_{global}$. $\alpha_l$ and $\alpha_g$ serve as the importance of $\textbf{z}_{l}$ and $\textbf{z}_{g}$ when computing the combined node embedding $\textbf{z}$. To better understand the contribution of $\textbf{z}_{l}$ and $\textbf{z}_{g}$, we conduct a detailed analysis of the learned attention weights $\alpha_l$ and $\alpha_g$ in the three tasks we experimented with. Since the node embeddings are directly related to the involved interactions, such analysis can also reveal the contribution of local and global interactions on the predictions in different tasks. In each task, we take an average of all $\alpha_l$ or $\alpha_g$ to be the overall importance of the corresponding group of interactions. Then we compare the computed average attention weights $\overline{\alpha_l}$ and $\overline{\alpha_g}$ and list the results in Table~\ref{table:attention}. A higher attention weight in each task indicates a stronger contribution of the corresponding interactions on solving the task. 

For the targets being predicted in QM9, we find that all of them have $\overline{\alpha_l} \geq \overline{\alpha_g}$ except the electronic spatial extent $\left\langle R^{2}\right\rangle$, indicating a stronger contribution of the local interactions, which are defined by chemical bonds in this task. This may be because QM9 contains small molecules with only up to 9 non-hydrogen atoms, local interactions can capture a considerable portion of all atomic interactions. However, when predicting electronic spatial extent $\left\langle R^{2}\right\rangle$, we notice that $\overline{\alpha_l} < \overline{\alpha_g}$, which suggests that $\left\langle R^{2}\right\rangle$ is mainly affected by the global interactions that are the pairwise interactions within $10\angstrom$ in this case. This is not surprising since $\left\langle R^{2}\right\rangle$ is the electric field area affected by the ions in the molecule, and is directly related to the diameter or radius of the molecule. Besides, previous study~\cite{garg2020generalization} has demonstrated that graph properties like diameter and radius cannot be computed by message passing-based GNNs that rely entirely on local information, and additional global information is needed. Thus it is expected that global interactions have a stronger contribution than local interactions on predicting electronic spatial extent.

For the RNA 3D structure prediction on RNA-Puzzles and the protein-ligand binding affinity prediction on PDBbind, we find $\overline{\alpha_l} < \overline{\alpha_g}$ in both cases, which indicates that global interactions play a more important role than local interactions. It is because the goals of these two tasks highly rely on global interactions, which are necessary for representing the global structure of RNA when predicting RNA 3D structure, and are crucial for capturing the relationships between protein and ligand when predicting binding affinity.

\begin{table}[t]
\small
\centering
\begin{tabular}{cccccccccccccc}
\toprule
    \multirow{2}{*}[-1.0ex]{\shortstack{Attention\\Weight}} & \multicolumn{11}{c}{QM9} & \multirow{2}{*}[-1.0ex]{\shortstack{RNA-\\Puzzles}} & \multirow{2}{*}[-1.0ex]{PDBbind}\\\cmidrule{2-12}
    & $\mu$ & $\alpha$ & $\epsilon_{\text{HOMO}}$ & $\epsilon_{\text{LUMO}}$ & $\left\langle R^{2}\right\rangle$ & ZPVE & $U_0$ & $U$ & $H$ & $G$ & $c_v$ & & \\
\midrule
    $\overline{\alpha_l}$ & \textbf{0.64} & \textbf{0.53} & 0.50 & 0.50 & 0.29 & \textbf{0.54} & \textbf{0.60} & \textbf{0.60} & \textbf{0.60} & \textbf{0.57} & \textbf{0.58} & 0.22 & 0.34 \\
    $\overline{\alpha_g}$ & 0.36 & 0.47 & 0.50 & 0.50 & \textbf{0.71} & 0.46 & 0.40 & 0.40 & 0.40 & 0.43 & 0.42 & \textbf{0.78} & \textbf{0.66} \\
\bottomrule
\end{tabular}
\caption{\textbf{Comparison of the average attention weights $\overline{\alpha_l}$ and $\overline{\alpha_g}$ for local and global interactions in attention pooling.} The higher attention weight for each target is marked in bold.}
\label{table:attention}
\end{table}

\section*{Conclusion}
In this work, we tackle the limitations of previous GNNs regarding their limited applicability and inefficiency for representation learning of molecular systems with 3D structures and propose a universal framework, \method, to accurately and efficiently learn the representations of 3D molecules in any molecular system. \method explicitly models local and non-local interaction as well as their combined effects inspired by molecular mechanics. The resulting framework incorporates rich geometric information like distances and angles when modeling local interactions, and avoids using expensive operations on modeling non-local interactions. Besides, \method learns the contribution of different interactions to combine the updated node embeddings for the final output. When designing the aforementioned operations in \method, we preserve E(3)-invariance for scalar output and preserve E(3)-equivariance for vectorial output to enable more applicable cases. In our experiments, we evaluate the performance of \method with state-of-the-art baselines on various tasks
involving different molecular systems, including small molecules, RNAs, and protein-ligand complexes. In each task, \method outperforms the corresponding baselines in terms of
both accuracy and efficiency. These results clearly demonstrate the generalization power of \method even though non-local interactions in molecules are modeled with only pairwise distances as geometric information. 

An under-investigated aspect of our proposed \method is that \method preserves E(3)-invariance in operations when predicting scalar properties while requiring additional representations and operations to preserve E(3)-equivariance for vectorial properties. Considering that various equivariant GNNs have been proposed for predicting either scalar or vectorial properties solely by preserving equivariance, it would be worth extending the idea in \method to equivariant GNNs with a potential to further improve both accuracy and efficiency. Another interesting direction is that although we only experiment \method on single-task learning, \method is promising to be used in multi-task learning across diverse tasks that involve molecules of varying sizes and types to gain better generalization. Besides using \method for predicting physiochemical properties of molecules, \method can be used as a universal building block for the representation learning of molecular systems in various molecular science problems. Another promising application of \method is self-supervised learning for molecular systems with few labeled data (e.g., RNA structures). For example, we can use the features in one graph layer to learn properties in another graph layer by utilizing the multiplex nature of \method.

\section*{Methods}
\subsection*{Details of \method}
In this section, we will describe \method in detail, including the involved features, embeddings, and operations.

\paragraph{Input features.}
The input features of \method include atomic features and geometric information as shown in Figure~\ref{fig:method}b. For atomic features, we use only atomic numbers $Z$ for the tasks on QM9 and RNA-Puzzles following~\cite{schutt2018schnetpack,unke2019physnet,klicpera_dimenet_2020,klicpera_dimenetpp_2020,townshend2021geometric}, and use 18 chemical features like atomic numbers, hybridization, aromaticity, partial charge, etc., for the task on PDBbind following~\cite{stepniewska2018development,li2021structure}. The atomic numbers $Z$ are represented by randomly initialized, trainable embeddings according to~\cite{schutt2018schnetpack,unke2019physnet,klicpera_dimenet_2020,klicpera_dimenetpp_2020}. For geometric information, we capture the needed pairwise distances and angles in the multiplex molecular graph $G$ as shown in Figure~\ref{fig:method}d. The features ($d$, $\theta$) for the distances and angles are computed with the basis functions in~\cite{klicpera_dimenet_2020} to reduce correlations. For the prediction of vectorial properties, we use the atomic position $\vec{r}$ to be the initial associated geometric vector $\vec{v}$ of each atom.

\paragraph{Message embeddings.}
In the message passing scheme~\cite{gilmer2017neural}, the update of node embeddings $\boldsymbol{h}$ relies on the passing of the related messages $\boldsymbol{m}$ between nodes. In \method, we define the input message embeddings $\boldsymbol{m}$ of message passing schemes with the following way:
\begin{align}
\boldsymbol{m}_{ji} &= \mathrm{MLP}_m([\boldsymbol{h}_{j} | \boldsymbol{h}_{i} | \boldsymbol{e}_{ji}]), 
\end{align}
where $i, j \in G_{global}$ or $G_{local}$ are connected nodes that can define a message embedding, $\mathrm{MLP}$ denotes the multi-layer perceptron, $|$ denotes the concatenation operation. The edge embedding $\boldsymbol{e}_{ji}$ encodes the corresponding pairwise distance $d$ between node $i, j$.

\paragraph{Global message passing.}
As depicted in Figure~\ref{fig:method}e, the Global Message Passing in each hidden layer of \method, which consists of a message block and an update block, updates the node embeddings $\boldsymbol{h}$ in $G_{global}$ by using the related adjacency matrix $\textbf{A}_{global}$ and pairwise distances $d_{global}$. The message block is defined as below to perform the message passing operation:
\begin{align}
\boldsymbol{h}_{i}^{t} &= \boldsymbol{h}_{i}^{t-1} + \sum\nolimits_{j \in \mathcal{N}(i)} \boldsymbol{m}_{ji}^{t-1}\odot \phi_{d}(\boldsymbol{e}_{j i}), \label{node_update_g}
\end{align}
where $i, j \in G_{global}$, $\phi_{d}$ is a learnable function, $\boldsymbol{e}_{ji}$ is the embedding of pairwise distance $d$ between node $i, j$, and $\odot$ denotes the element-wise production. After the message block, an update block is used to compute the node embeddings $\boldsymbol{h}$ for the next layer as well as the output $\textbf{z}$ for this layer. We define the update block using a stack of three residual blocks, where each residual block consists of a two-layer MLP and a skip connection across the MLP. There is also a skip connection between the input of the message block and the output of the first residual block. After the residual blocks, the updated node embeddings $\boldsymbol{h}$ are passed to the next layer. For the output $\textbf{z}$ of this layer to be combined in the fusion module, we further use a three-layer MLP to get $\textbf{z}$ with desired dimension size.

\paragraph{Local message passing.} 
For the updates of node embeddings $\boldsymbol{h}$ in $G_{local}$, we incorporate both pairwise distances $d_{local}$ and angles $\theta_{local}$ as shown in Figure~\ref{fig:method}e. To capture $\theta_{local}$, we consider up to the two-hop neighbors of each node. In Figure~\ref{fig:method}d, we show an example of the angles we considered: Some angles are between one-hop edges and two-hop edges (e.g. $\angle i j_1 k_1$), while the other angles are between one-hop edges (e.g. $\angle j_1 i j_2$). Compared to previous GNNs~\cite{klicpera_dimenet_2020,klicpera_dimenetpp_2020,shui2020heterogeneous} that incorporate only part of these angles, our \method is able to encode the geometric information more comprehensively. In the Local Message Passing, we also use a message block and an update block following the design of the Global Message Passing as shown in Figure~\ref{fig:method}e. However, the message block is defined differently as the one in the Global Message Passing to encode additional angular information:
\begin{align}
\boldsymbol{m}_{ji}^{'t-1} &= \boldsymbol{m}_{ji}^{t-1} + \sum_{j' \in \mathcal{N}(i)\setminus\{j\}} \boldsymbol{m}_{j'i}^{t-1} \odot \phi_{d}(\boldsymbol{e}_{j'i}) \odot \phi_{\theta}(\boldsymbol{a}_{j'i, j i}) 
+ \sum_{k \in \mathcal{N}(j)\setminus\{i\}} \boldsymbol{m}_{kj}^{t-1} \odot \phi_{d}(\boldsymbol{e}_{kj}) \odot \phi_{\theta}(\boldsymbol{a}_{k j, j i}), \label{message_update} \\
\boldsymbol{h}_{i}^{t} &= \boldsymbol{h}_{i}^{t-1} + \sum_{j \in \mathcal{N}(i)} \boldsymbol{m}_{ji}^{'t-1}\odot \phi_{d}(\boldsymbol{e}_{ji}), \label{node_update_l}
\end{align}
where $i, j, k \in G_{local}$, $\boldsymbol{e}_{ji}$ is the embedding of pairwise distance $d$ between node $i, j$, $\boldsymbol{a}_{k j, j i}$ is the embedding of angle $\theta_{k j, j i}=\angle kji$ defined by node $i, j, k$, and $\phi_{d}, \phi_{\theta}$ are learnable functions. In Equation (\ref{message_update}), we use two summation terms to separately encode the angles in different hops with the associated pairwise distances to update $\boldsymbol{m}_{ji}$. Then in Equation (\ref{node_update_l}), the updated message embeddings $\boldsymbol{m}_{ji}'$ are used to perform message passing. After the message block, we use the same update block as the one used in the Global Message Passing for updating the learned node embeddings.

\paragraph{Fusion module.}
The fusion module consists of two steps of pooling as shown in Figure~\ref{fig:method}b. In the first step, attention pooing is utilized to learn the combined embedding $\textbf{z}^{t}$ based on the output node embeddings $\textbf{z}_{g}^{t}$ and $\textbf{z}_{l}^{t}$ in each hidden layer $t$. The detailed architecture of attention pooling is illustrated in Figure~\ref{fig:method}e. We first compute the attention weight $\alpha_{\textcolor{black}{p},i}$ on node $i$ that measures the contribution of the results from \textcolor{black}{plex or} graph layer $\textcolor{black}{p} \in \{g, l\}$ in multiplex graph $G$:
\begin{align}
\alpha_{\textcolor{black}{p},i}^{t} = \frac{\exp(\operatorname {LeakyReLU} (\boldsymbol{W}_{\textcolor{black}{p}}^t \textbf{z}_{\textcolor{black}{p},i}^t))} {\sum_{\textcolor{black}{p}}\exp( \operatorname{LeakyReLU}(\boldsymbol{W}_{\textcolor{black}{p}}^t \textbf{z}_{\textcolor{black}{p},i}^t)) },\label{softmax}
\end{align}
where $\boldsymbol{W}_{\textcolor{black}{p}}^t \in \mathbb{R}^{1\times F}$ is a learnable weight matrix different for each hidden layer $t$ and graph layer $\textcolor{black}{p}$, and $F$ is the dimension size of $\textbf{z}_{\textcolor{black}{p},i}^t$. With $\alpha_{\textcolor{black}{p},i}^{t}$, we can compute the combined node embedding $\textbf{z}_i^t$ of node $i$ using a weighted summation:
\begin{align}
\textbf{z}_i^t = \sum\nolimits_{\textcolor{black}{p}} \alpha_{\textcolor{black}{p},i}^{t} (\boldsymbol{W}_{\textcolor{black}{p}}^{'t}\textbf{z}_{\textcolor{black}{p},i}^t), \label{weight_sum}
\end{align}
where $\boldsymbol{W}_{\textcolor{black}{p}}^{'t} \in \mathbb{R}^{D\times F}$ is a learnable weight matrix different for each hidden layer $t$ and graph layer $\textcolor{black}{p}$, $D$ is the dimension size of $\textbf{z}_i^t$, and $F$ is the dimension size of $\textbf{z}_{\textcolor{black}{p},i}^t$. 

In the second step of the fusion module, we sum the combined node embedding $\textbf{z}$ of all hidden layers to compute the final node embeddings $\boldsymbol{y}$. If a graph-level embedding $\boldsymbol{y}$ is desired, we compute as follows: 
\begin{align}
\boldsymbol{y} = \sum\nolimits_{i=1}^{N}\sum\nolimits_{t=1}^{T} \textbf{z}_i^t.\label{sum}
\end{align}

\paragraph{Preservation of E(3)-invariance \& E(3)-equivariance.}
For the operations described above, they preserve the E(3)-invariance of the input atomic features and geometric information and can predict E(3)-invariant scalar properties. To predict E(3)-equivariant vectorial property $\vec{u}$, we introduce an associated geometric vector $\vec{v}_i$ for each node $i$ and extend \method to preserve the E(3)-equivariance for learning $\vec{u}$. In detail, the associated geometric vector $\vec{v}_i^{t}$ of node $i$ in hidden layer $t$ is defined as:
\begin{align}
\vec{v}_i^{t} = f_v(\{\boldsymbol{h}^{t}\}, \{\vec{r}\}),\label{vector}
\end{align}
where $\{\boldsymbol{h}^{t}\}$ denotes the set of learned node embeddings of all nodes in hidden layer $t$, $\{\vec{r}\}$ denotes the set of position vectors of all nodes in 3d coordinate space, and $f_v$ is a function that preserves the E(3)-equivariance of $\vec{v}_i^{t}$ with respect to $\{\vec{r}\}$. Equation (\ref{vector}) is computed after each message passing module in \method.

To predict a final vectorial property $\vec{u}$, we modify Equation (\ref{weight_sum}) and (\ref{sum}) in the fusion module as the following operations:
\begin{align}
\vec{u}_{i}^{t} &= \sum\nolimits_{\textcolor{black}{p}} \alpha_{\textcolor{black}{p},i}^{t} (\boldsymbol{W}_{\textcolor{black}{p}}^{'t}\textbf{z}_{\textcolor{black}{p},i}^t) \vec{v}_{\textcolor{black}{p},i}^t,\label{weight_sum_vec} \\
\vec{u} &= \sum\nolimits_{i=1}^{N}\sum\nolimits_{t=1}^{T} \vec{u}_{i}^{t},\label{sum_vec}
\end{align}
where $\vec{v}_{\textcolor{black}{p},i}^t$ is the associated geometric vector of node $i$ on graph layer $\textcolor{black}{p}$ in hidden layer $t$, $\vec{u}_{i}^{t}$ is the learned vector of node $i$ in hidden layer $t$, and $\boldsymbol{W}_{\textcolor{black}{p}}^{'t} \in \mathbb{R}^{1\times F}$ is a learnable weight matrix different for each hidden layer $t$ and graph layer $\textcolor{black}{p}$. In Equation (\ref{weight_sum_vec}), we multiply $\vec{v}_{\textcolor{black}{p},i}^t$ with the learned scalar node contributions. In Equation (\ref{sum_vec}), we sum all node-level vectors in all hidden layers to compute the final prediction $\vec{u}$.

For predicting dipole moment $\vec{\mu}$ , \textcolor{black}{which is an E(3)-equivariant vectorial property that describes the net molecular polarity in electric field}, we design $f_v$ in Equation (\ref{vector}) as motivated by quantum mechanics ~\cite{veit2020predicting}. \textcolor{black}{The conventional method to compute molecular dipole moment involves approximating electronic charge densities as concentrated at each atomic position, resulting in $\vec{\mu}=\sum\nolimits_{i}\vec{r}_{c,i}q_i$, where $q_i$ is the partial charge of node $i$, and $\vec{r}_{c,i}=\vec{r}_i - (\sum\nolimits_{i}\vec{r}_i)/N$ is the relative atomic position of node $i$. However, this approximation is not accurate enough. Instead, we use a more accurate} approximation by adding dipoles onto atomic positions in the distributed multipole analysis (DMA) approach~\cite{stone1981distributed}. This results in the dipole moment equation: $\vec{\mu}=\sum\nolimits_{i}(\vec{r}_{c,i}q_i+\vec{\mu}_i)$, where $\vec{\mu}_i$ is the associated partial dipole of node $i$. The equation can be rewritten as $\vec{\mu}=\sum\nolimits_{i}f_v(\vec{r}_{i})q_i$, where $q_i$ is the scalar atomic contribution that can be modeled by an invariant fashion. By treating $f_v(\vec{r}_{i})$ as $\vec{v}_{i}^t$, the equation has a similar format as a combination of Equation (\ref{weight_sum_vec}) and Equation (\ref{sum_vec}). We update $\vec{v}_{i}^t$ in the following way:  
\begin{align}
\vec{v}_{i}^{t} =\sum\nolimits_{j \in \mathcal{N}(i)}(\vec{r}_{i} - \vec{r}_{j})\lVert \boldsymbol{m}_{i j}^{t}\rVert,
\end{align}
where $\lVert \cdot \rVert$ denotes the L2 norm. Since $\vec{v}_{i}^{t}$ as well as $\vec{\mu}$ are computed by a linear combination of $\{\vec{r}\}$, our \method can preserve E(3)-equivariance with respect to $\{\vec{r}\}$ when performing the prediction.

\paragraph{Computational Complexity.}
We analyze the computational complexity of \method by addressing the number of messages. We denote the cutoff distance when creating the edges as $d_g$ and $d_l$ in $G_g$ and $G_l$. The average degree is $k_g$ in $G_g$ and is $k_l$ in $G_l$. In each hidden layer of \method, Global Message Passing needs $O(Nk_g)$ messages because it requires one message for each pairwise distance between the central node and its one-hop neighbor. While Local Message Passing requires one message for each one-hop or two-hop angle around the central node. The number of angles can be estimated as follows: For $k$ edges connected to a node, they can define $(k(k-1))/2$ angles which result in a complexity of $O(Nk^2)$. The number of one-hop angles and two-hop angles all has such complexity. So that Local Message Passing needs $O(2N{k_l}^2)$ messages. In total, \method requires the computation of $O(Nk_g+2N{k_l}^2)$ messages in each hidden layer, while previous approaches~\cite{klicpera_dimenet_2020,klicpera_dimenetpp_2020, shui2020heterogeneous,li2021structure,liu2022spherical} require $O(N{k_g}^2)$ messages. For 3D molecules, we have $k_g \propto {d_g}^3$ and $k_l \propto {d_l}^3$. With proper choices of $d_l$ and $d_g$, we have $k_l \ll k_g$. In such cases, our model is more efficient than the related GNNs. We here list the comparison of the number of messages needed in our experiments as an example: On QM9 with $d_g=5\angstrom$, our model needs 0.5k messages/molecule on average, while DimeNet++ needs 4.3k messages. On PDBBind with $d_l=2\angstrom$ and $d_g=6\angstrom$, our model needs only 12k messages/molecule on average, while DimeNet++ needs 264k messages.

\subsection*{Data collection and processing}

\paragraph{QM9.}
For QM9, we use the source provided by~\cite{ramakrishnan2014quantum}. Following the previous works~\cite{klicpera_dimenet_2020,klicpera_dimenetpp_2020, shui2020heterogeneous}, we process QM9 by removing about 3k molecules that fail a geometric consistency check or are difficult to converge~\cite{faber2017prediction}. For properties $U_0$, $U$, $H$, and $G$, only the atomization energies are used by subtracting the atomic reference energies as in~\cite{klicpera_dimenet_2020,klicpera_dimenetpp_2020, shui2020heterogeneous}. For property $\Delta \epsilon$, we follow the same way as the DFT calculation and predict it by calculating $\epsilon_{\mathrm{LUMO}}-\epsilon_{\mathrm{HOMO}}$. For property $\mu$, the final result is the magnitude of the predicted vectorial $\boldsymbol{\mu}$ when using our geometric vector-based approaches with \method. The 3D molecular structures are processed using the RDKit library~\cite{landrum2013rdkit}. Following~\cite{klicpera_dimenet_2020}, we randomly use 110000 molecules for training, 10000 for validation and 10831 for testing. In our multiplex molecular graphs, we use chemical bonds as the edges in the local layer, and a cutoff distance (5 or 10$\angstrom$) to create the edges in the global layer.

\paragraph{RNA-Puzzles.} RNA-Puzzles consists of the first 21 RNAs in the RNA-Puzzles structure prediction challenge~\cite{miao2020rna}. Each RNA is used to generate at least 1500 structural models using FARFAR2, where 1$\%$ of the models are near native (i.e., within a 2$\angstrom$ RMSD of the experimentally determined native structure). Following~\cite{townshend2021geometric}, we only use the carbon, nitrogen, and oxygen atoms in RNA structures. When building multiplex graphs for RNA structures, we use cutoff distance $d_l=2.6\angstrom$ for the local interactions in $G_{local}$ and $d_g=20\angstrom$ for the global interactions in $G_{global}$. 

\paragraph{PDBBind.} For PDBBind, we use PDBbind v2016 following~\cite{stepniewska2018development,li2021structure}. Besides, we use the same data splitting method according to~\cite{li2021structure} for a fair comparison. In detail, we use the core subset which contains 290 complexes in PDBbind v2016 for testing. The difference between the refined and core subsets, which includes 3767 complexes, is split with a ratio of 9:1 for training and validation. We use log$K_i$ as the target property being predicted, which is proportional to the binding free energy. In each complex, we exclude the protein residues that are more than 6$\angstrom$ from the ligand and remove all hydrogen atoms. The resulting complexes contain around 300 atoms on average. In our multiplex molecular graphs, we use cutoff distance $d_l=2\angstrom$ in the local layer and $d_g=6\angstrom$ in the global layer.

\subsection*{Experimental settings}
In our message passing operations, we define $\phi_{d}(\boldsymbol{e})=\boldsymbol{W}_{\boldsymbol{e}}\boldsymbol{e}$ and $\phi_{\alpha}(\boldsymbol{\alpha})=\mathrm{MLP}_{\alpha}(\boldsymbol{\alpha})$, where $\boldsymbol{W}_{\boldsymbol{e}}$ is a weight matrix, $\mathrm{MLP}_{\alpha}$ is a multi-layer perceptron (MLP). All MLPs used in our model have two layers by taking advantage of the approximation capability of MLP~\cite{hornik1989multilayer}. For all activation functions, we use the self-gated Swish activation function~\cite{ramachandran2017searching}. For the basis functions, we use the same parameters as in~\cite{klicpera_dimenet_2020}. To initialize all learnable parameters, we use the default settings used in PyTorch without assigning specific initializations except the initialization for the input node embeddings on QM9: $\boldsymbol{h}$ are initialized with random values uniformly distributed between $-\sqrt{3}$ and $\sqrt{3}$. 
In all experiments, we use the Adam optimizer~\cite{kingma2014adam} to minimize the loss. In Supplementary Table~\ref{table:hyperparameter}, we list the typical hyperparameters used in our experiments. All of the experiments are done on an NVIDIA Tesla V100 GPU (32 GB).

\paragraph{Small molecule property prediction.}
In our experiment on QM9, we use the single-target training following~\cite{klicpera_dimenet_2020} by using a separate model for each target instead of training a single shared model for all targets. The models are optimized by minimizing the mean absolute error (MAE) loss. We use a linear learning rate warm-up over 1 epoch and an exponential decay with a ratio 0.1 every 600 epochs. The model parameter values for validation and testing are kept using an exponential moving average with a decay rate of 0.999. To prevent overfitting, we use early stopping on the validation loss. For properties ZPVE, $U_0$, $U$, $H$, and $G$, we use the cutoff distance in the global layer $d_g=5\angstrom$. For the other properties, we use $d_g=10\angstrom$. We repeat our runs 3 times for each \method variant following~\cite{anderson2019cormorant}. 

\paragraph{RNA 3D structure prediction.}
\method is optimized by minimizing the smooth L1 loss\cite{ren2015faster} between the predicted value and the ground truth. An early-stopping strategy is adopted to decide the best epoch based on the validation loss.

\paragraph{Protein-ligand binding affinity prediction.}
We create three weight-sharing, replica networks, one each for predicting the target $G$ of complex, protein pocket, and ligand following~\cite{gomes2017atomic}. The final target is computed by $\Delta G_{\text{complex}} = G_{\text{complex}} - G_{\text{pocket}} - G_{\text{ligand}}$. The full model is trained by minimizing the mean absolute error (MAE) loss between $\Delta G_{\text{complex}}$ and the true values. The learning rate is dropped by a factor of 0.2 every 50 epochs. Moreover, we perform 5 independent runs according to~\cite{li2021structure}.

\paragraph{Efficiency comparison. }
In the experiment on investigating the efficiency of \method, we use NVIDIA Tesla V100 GPU (32 GB) for a fair comparison. For small molecule property prediction, we use the related models for predicting property $U_0$ of QM9 as an example. We use batch size=128 for all models and use the configurations reported in the corresponding papers. For RNA 3D structure prediction, we use \method and ARES to predict the structural models of RNA in puzzle 5 of RNA-Puzzles challenge. The RNA being predicted has 6034 non-hydrogen atoms. The model settings of \method and ARES are the same as those used for reproducing the best results. We use batch size=8 when performing the predictions. For protein-ligand binding affinity prediction, we use the configurations that can reproduce the best results for the related models.

\section*{Data Availability}
The QM9 dataset is available at \url{https://figshare.com/collections/Quantum_chemistry_structures_and_properties_of_134_kilo_molecules/978904}. The datasets for RNA 3D structure prediction can be found at \url{https://purl.stanford.edu/bn398fc4306}. The PDBbind v2016 dataset is available at \url{http://www.pdbbind.org.cn/} or \url{https://github.com/PaddlePaddle/PaddleHelix/tree/dev/apps/drug_target_interaction/sign}.

\section*{Code Availability}
The source code of our model is publicly available on GitHub at the following repository: \url{https://github.com/XieResearchGroup/Physics-aware-Multiplex-GNN}.

\bibliography{main}

\section*{Acknowledgements}
This project has been funded with federal funds from the National Institute of General Medical Sciences of National Institute of Health (R01GM122845) and the National Institute on Aging of the National Institute of Health (R01AG057555).

\section*{Author Contributions}
L.X. and S.Z. conceived and designed the method and the experiments.  S.Z. and Y.L. prepared the data. S.Z. implemented the algorithm and performed the experiments. All authors wrote and reviewed the manuscript.

\section*{Additional Information}
\subsection*{Competing interests}
The authors declare no competing interests.

\newpage
\appendix

\setcounter{figure}{0}
\renewcommand{\thefigure}{S\arabic{figure}}

\setcounter{table}{0}
\renewcommand{\thetable}{S\arabic{table}}

\section*{Appendix}

\subsection*{Details of baselines}
The following methods are being compared with our PAMNet in experiments:

\paragraph{Small molecule property prediction}
\begin{itemize}[leftmargin=10pt]
\item \textbf{SchNet}~\cite{schutt2018schnetpack} is a GNN that uses continuous-filter convolutional layers to model atomistic systems. Interatomic distances are used when designing convolutions.
\item \textbf{PhysNet}~\cite{unke2019physnet} uses message passing scheme for predicting properties of chemical systems. It models chemical interactions with learnable distance-based functions.
\item \textbf{MGCN}~\cite{lu2019molecular} utilizes the multilevel structure in molecular system to learn the representations of quantum interactions level by level based on GNN. The final molecular property prediction is made with the overall interaction representation.
\item \textbf{PaiNN}~\cite{schutt2021equivariant} is a GNN that augments the invariant SchNet into equivariant flavor by projecting the pairwise distances via radial basis functions and iteratively updates the geometric vectors along with the scalar features.
\item \textbf{DimeNet++}~\cite{klicpera_dimenetpp_2020} is an improved version of DimeNet~\cite{klicpera_dimenet_2020} with better accuracy and faster speed. It can also be used for non-equilibrium molecular structures.
\item \textbf{SphereNet}~\cite{liu2022spherical} is a GNN method that achieves local completeness by incorporating comprehensive 3D information like distance, angle, and torsion information for 3D graphs.
\end{itemize}

\paragraph{RNA 3D structure prediction}
\begin{itemize}[leftmargin=10pt]
\item \textbf{ARES}~\cite{townshend2021geometric} is a state-of-the-art machine learning approach for identifying accurate RNA 3D structural models from candidate ones. It is a GNN that integrates rotational equivariance into the message passing.
\item \textbf{Rosetta}~\cite{watkins2020farfar2} is a molecular modeling software package that provides tools for RNA 3D structure prediction.
\item \textbf{RASP}~\cite{capriotti2011all} is a full-atom knowledge-based potential with geometrical descriptors for RNA structure prediction.
\item \textbf{3dRNAscore}~\cite{wang20153drnascore} is an all-heavy-atom knowledge-based potential that combines distance-dependent and dihedral-dependent energies for identifying native RNA structures and ranking predicted structures. 
\end{itemize}

\paragraph{Protein-ligand binding affinity prediction}
\begin{itemize}[leftmargin=10pt]
\item \textbf{ML-based methods} include linear regression (LR), support vector regression (SVR), and random forest (RF). These approaches use the inter-molecular interaction features introduced in RF-Score~\cite{ballester2010machine} as input for prediction.
\item \textbf{Pafnucy}~\cite{stepniewska2018development} is a representative 3D CNN-based model that learns the spatial structure of protein-ligand complexes.
\item \textbf{OnionNet}~\cite{zheng2019onionnet} is a CNN-based method that generates 2D interaction features by considering rotation-free element-pair contacts in complexes.
\item \textbf{GraphDTA}~\cite{nguyen2021graphdta} uses GNN models to learn the complex graph and utilizes CNN to learn the protein sequence. We use the best-performed variant (GAT-GCN) for comparison.
\item \textbf{SGCN}~\cite{danel2020spatial} utilizes atomic coordinates and leverages node positions based on graph convolutional network~\cite{kipf2016semi}.
\item \textbf{GNN-DTI}~\cite{lim2019predicting} is a distance-aware graph attention network~\cite{velivckovic2018graph} that considers 3D structural information to learn the intermolecular interactions in protein-ligand complexes.
\item \textbf{D-MPNN}~\cite{yang2019analyzing} is a message passing neural network that incorporates edge features. The aggregation process addresses the pairwise distance information contained in edge features.
\item \textbf{MAT}~\cite{maziarka2020molecule} utilizes inter-atomic distances and employs a molecule-augmented attention mechanism based on transformers for graph representation learning.
\item \textbf{DimeNet}~\cite{klicpera_dimenet_2020} is a message passing neural network using directional message passing scheme for small molecules. Both distances and angles are used when modeling the molecular interactions.
\item \textbf{CMPNN}~\cite{song2020communicative} is built based on D-MPNN and has a communicative message passing scheme between nodes and edges for better performance when learning molecular representations.
\item \textbf{SIGN}~\cite{li2021structure} is a recent state-of-the-art GNN for predicting protein-ligand binding affinity. It builds complex interaction graphs for protein-ligand complexes and integrates both distance and angle information in modeling.
\end{itemize}

For small molecule property prediction, we use the baseline results reported in their original works for baselines. For RNA 3D structure prediction, we use the baseline results in~\cite{townshend2021geometric}. For protein-ligand binding affinity prediction, we use the baseline results in~\cite{li2021structure}. When performing efficiency evaluation in our experiments, we adopt the public-available implementations of the related models: For DimeNet and DimeNet++, we adopt the implementation by PyTorch Geometric~\cite{FeyLenssen2019} at \url{https://github.com/rusty1s/pytorch_geometric/blob/73cfaf7e09/examples/qm9_dimenet.py}. For SphereNet, we use the official implementation at \url{https://github.com/divelab/DIG}. For ARES, we use the official implementation at \url{https://zenodo.org/record/6893040}. For SIGN, we use the official implementation at \url{https://github.com/PaddlePaddle/PaddleHelix/tree/dev/apps/drug_target_interaction/sign}.

\subsection*{Detailed analysis of near-native ranking task on RNA-Puzzles}  \label{rna_appendix}

\begin{figure*}[ht]
    \centering
    \includegraphics[width=0.7\columnwidth]{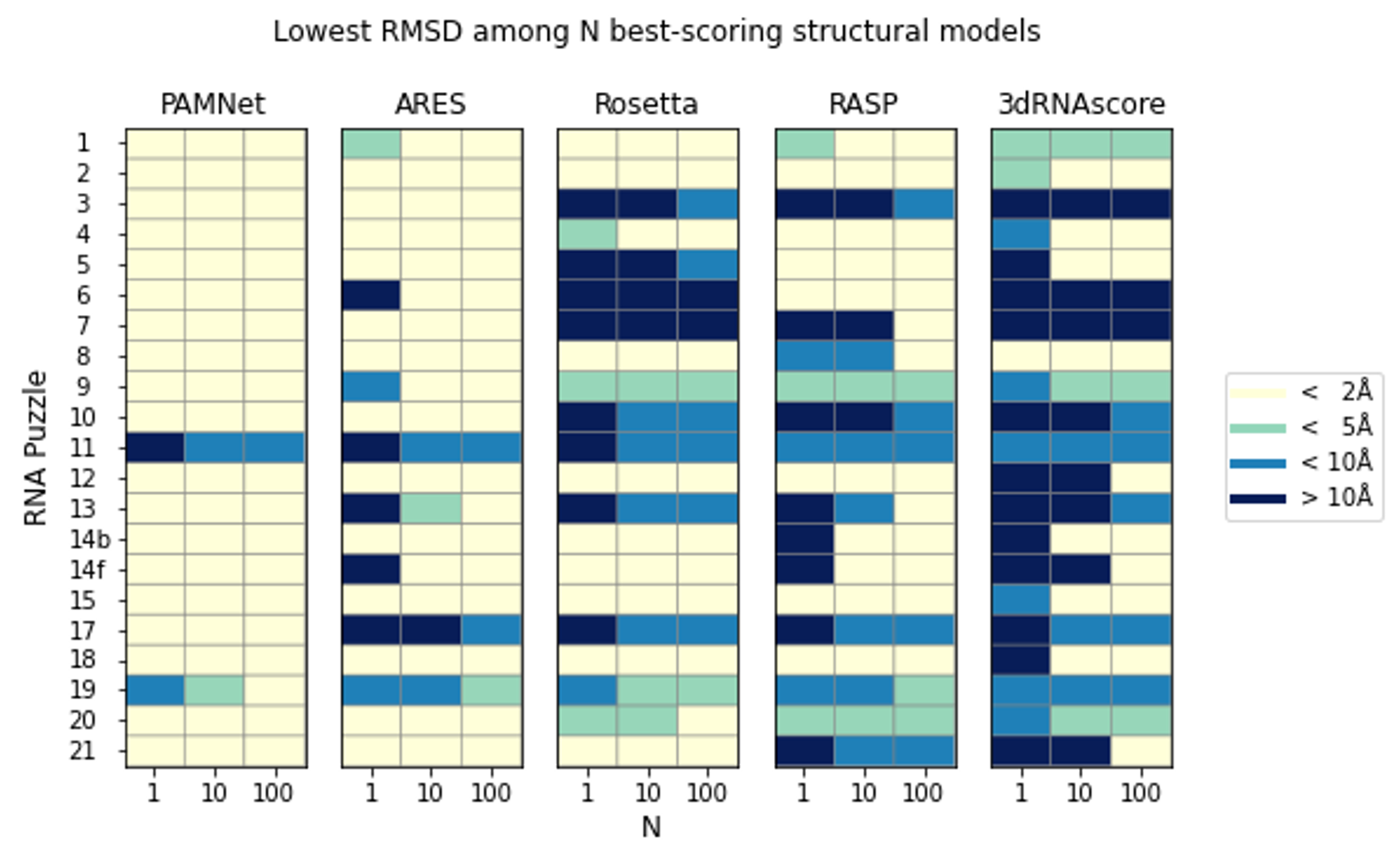}
    \caption{\textbf{Detailed analysis of near-native ranking task on RNA-Puzzles.} The results of the lowest RMSD among N best-scoring structural models of each RNA predicted by each scoring function are compared.}
\label{fig:rna_appendix}
\end{figure*}

For each RNA in RNA-Puzzles, we rank the structural models using \method and four baseline scoring functions. For each scoring function, we select the N $\in \{1, 10, 100\}$ best-scoring structural models for each RNA. For each RNA, scoring function, and N, we show the lowest RMSD across structural models in Figure~\ref{fig:rna_appendix}). The RMSD results are quantized to determine if each RMSD is below 2$\angstrom$, between 2$\angstrom$ and 5$\angstrom$, between 5$\angstrom$ and 10$\angstrom$, or above 10$\angstrom$. From the results, we find that for each RMSD threshold (2$\angstrom$, 5$\angstrom$, or 10$\angstrom$) and for each N, the number of RNAs with at least one selected model that has RMSD below the threshold is greater when using \method than when using any of the other four baseline scoring functions.

\subsection*{Statistical significance between SIGN and \method on PDBbind}
We use p-value to compute the statistical significance between SIGN and \method on PDBbind. As shown in Table~\ref{table:significance}, \method performs significantly better than SIGN on all four metrics with p-value < 0.05.

\subsection*{Detailed results of ablation study on QM9}
In Table~\ref{table:ablation_qm9}, we list the results of all properties on QM9 in our ablation study.

\begin{table}[ht]
\centering
\begin{tabular}{ccccc}
    \toprule
    Model & RMSE $\downarrow$ & MAE $\downarrow$ & SD $\downarrow$ & R $\uparrow$ \\
    \midrule
    SIGN & 1.316 (0.031) & 1.027 (0.025) & 1.312 (0.035) & 0.797 (0.012)\\
    \textbf{\method} & \textbf{1.263 (0.017)} & \textbf{0.987 (0.013)} & \textbf{1.261 (0.015)} & \textbf{0.815 (0.005)}\\
    \midrule
    Significance (p-value) & 0.0122 & 0.0156 & 0.0242 & 0.0212\\
    \bottomrule
\end{tabular}
\caption{\textbf{Statistical significance (p-value) between \method and SIGN on PDBbind.} The best results are marked in bold.}
\label{table:significance}
\end{table}

\begin{table}[ht]
\small
\centering
\begin{tabular}{lcccccccccccc}
\toprule
     Model & $\mu$ & $\alpha$ & $\epsilon_{\text{HOMO}}$ & $\epsilon_{\text{LUMO}}$ & $\delta \epsilon$ & $\left\langle R^{2}\right\rangle$ & ZPVE & $U_0$ & $U$ & $H$ & $G$ & $c_v$ \\
     \midrule
     PAMNet & \textbf{10.8} & \textbf{0.0447} & \textbf{22.8} & \textbf{19.2} & \textbf{31.0} & \textbf{0.093} & \textbf{1.17} & \textbf{5.90} & \textbf{5.92} & \textbf{6.04} & \textbf{7.14} & \textbf{0.0231} \\
     PAMNet w/o Attention Pooling & 11.1 & 0.0469 & 24.2 & 20.3 & 32.8 & 0.094 & 1.22 & 6.12 & 6.15 & 6.29 & 7.44 & 0.0234 \\
     PAMNet w/o Local MP & 13.9 & 0.0512 & 27.8 & 23.3 & 37.6 & 0.104 & 1.27 & 7.55 & 7.57 & 7.74 & 9.13 & 0.0262 \\
     PAMNet w/o Global MP & 21.8 & 0.0887 & 41.5 & 34.9 & 56.4 & 5.53 & 1.52 & 8.80 & 8.81 & 9.01 & 10.6 & 0.0316 \\
\bottomrule
\end{tabular}
\caption{\textbf{Results of all properties on QM9 in ablation study.}}
\label{table:ablation_qm9}
\end{table}

\begin{table}[ht]
\centering
\begin{tabular}{lccc}
    \toprule
    \multirow{2}{*}{Hyperparameters} & \multicolumn{3}{c}{Value}\\
      & QM9 & RNA-Puzzles & PDBbind\\
    \midrule
    Batch Size & 32, 128 & 8 & 32\\
    Hidden Dim. & 128 & 16 & 128\\
    Initial Learning Rate & 1e-4 & 1e-4 & 1e-3\\
    Number of Layers & 6 & 1 & 3\\
    Max. Number of Epochs & 900 & 50 & 100\\
    \bottomrule
\end{tabular}
\caption{\textbf{List of typical hyperparameters used in our experiments.}}
\label{table:hyperparameter}
\end{table}

\end{document}